\documentclass[10pt]{article} 
\usepackage{rlc_for_arXiv}

\usepackage{amssymb}            
\usepackage{mathtools}          
\usepackage{mathrsfs}           
\mathtoolsset{showonlyrefs}     
\usepackage{graphicx}           
\usepackage{subcaption}         
\usepackage[space]{grffile}     
\usepackage{url}                

\usepackage{amsmath,amsfonts}
\usepackage{algorithmic}
\usepackage{algorithm}
\usepackage{array}
\usepackage{textcomp}
\usepackage{stfloats}
\usepackage{verbatim}
\usepackage{times}
\usepackage{soul}

\usepackage{booktabs}
\usepackage{xcolor}
\usepackage[switch]{lineno}
\usepackage{bbm}

\usepackage{tikz}
\usetikzlibrary{positioning}
\usepackage{multirow}

\usepackage{caption}
\usepackage{subcaption}
\usepackage{epstopdf}

\title{Exploring the Link Between Bayesian Inference and\\
 Embodied Intelligence: Toward Open Physical-World \\ Embodied AI Systems}

\author{Bin Liu \\
    bins@ieee.org\\
    Home Robotics Lab, E-surfing Digital Life Technology Co., Ltd., China Telecom
}

\begin{document}

\maketitle

\begin{abstract}
Embodied intelligence posits that cognitive capabilities fundamentally emerge from - \emph{and are shaped by} - an agent's real-time sensorimotor interactions with its environment. Such adaptive behavior inherently requires continuous inference under uncertainty. Bayesian statistics offers a principled probabilistic framework to address this challenge by representing knowledge as probability distributions and updating beliefs in response to new evidence. The core computational processes underlying embodied intelligence - including perception, action selection, learning, and even higher-level cognition - can be effectively understood and modeled as forms of Bayesian inference.

Despite the deep conceptual connection between Bayesian statistics and embodied intelligence, Bayesian principles have not been widely or explicitly applied in today’s embodied intelligence systems. In this work, we examine both Bayesian and contemporary embodied intelligence approaches through two fundamental lenses: \textbf{search} and \textbf{learning} - the two central themes in modern AI, as highlighted in Rich Sutton's influential essay ``The Bitter Lesson". This analysis sheds light on why Bayesian inference has not played a central role in the development of modern embodied intelligence. At the same time, it reveals that current embodied intelligence systems remain largely confined to closed-physical-world environments, and highlights the potential for Bayesian methods to play a key role in extending these systems toward truly open physical-world embodied intelligence.

\textbf{\emph{Keywords}}: Embodied Intelligence, Bayesian Inference, the Bitter Lesson, Learning, Search, LLM, VLA, AI Agent, World Model, Foundation Model, Robotics, Optimization
\end{abstract}

\section{Introduction}
\label{sec:intro}
Embodied intelligence, a rapidly emerging subfield of AI, is gaining increasing attention from both academia and industry. Closely related concepts include pre-trained large language models (LLMs) \cite{achiam2023gpt,touvron2023llama,bai2023qwen,team2024qwen2,liu2024deepseek,glm2024chatglm}, reinforcement learning \cite{sutton1998reinforcement,luo2024precise,franccois2018introduction}, AI agents \cite{zhou2023webarena,wang2024survey,xi2025rise,deng2025ai,liu2025advances,li2023camel,wang2023describe,huang2022language,hu2023enabling,zhou2023large,wang2023voyager,fan2024videoagent,liu2023reason,liu2023agentbench,zhang2023exploring,zhai2024fine,deng2023mind2web,fan2022minedojo,shi2017world,zheng2024agentstudio,huang2023embodied,wang2024jarvis}, spatial intelligence \cite{yang2025thinking,mao2025spatiallm,huang2024rekep}, 3D vision \cite{agia2022taskography,rana2023sayplan,zhen20243d,xu2024embodiedsam,gao2023visfusion,sun2021neuralrecon,lee2024guess}, world models \cite{zhen20243d,xiang2023language,mazzaglia2024genrl,nottingham2023embodied,wang2025founder,zhou2024robodreamer,gupta2024essential}, and multimodal models \cite{driess2023palm,li2022blip,zhang2023multimodal,radford2021learning,xing2019adaptive,wu2020visual,lee2019making,zhang2023prompt,yu2024boosting,ramesh2021zero,garcia2024towards,yang2025magma,fan2024videoagent}, especially visual-language-action (VLA) models \cite{zhen20243d,black2410pi0,zitkovich2023rt,intelligence2025pi_,zhong2025dexgraspvla,pan2024vision,liu2024robomamba,zhang2024hirt,brohan2022rt,zitkovich2023rt,vuong2023open,belkhale2024rt,gu2023rt,leal2024sara,ahn2024autort,bu2025agibot,liu2024rdt,kim2024openvla,bjorck2025gr00t,cheang2024gr,li2024cogact}.

Bayesian statistics, particularly Bayesian inference, represents one of the two major schools of thought in statistical science - the other being the frequentist school \cite{berger2013statistical,gelman2013philosophy,box2011bayesian,tipping2003bayesian,glickman2007basic,kass1995bayes}. While Bayesian methods may seem somewhat traditional compared to these contemporary AI approaches, there is a deep, intrinsic conceptual connection between Bayesian inference and embodied intelligence.

Bayesian approaches emphasize that an agent’s understanding of the world is inherently uncertain, representing beliefs probabilistically and continuously updating prior knowledge in light of new evidence \cite{korb2010bayesian,ghahramani2015probabilistic}. Similarly, embodied intelligence holds that an agent’s cognitive capabilities fundamentally emerge from - \emph{and are shaped by} - its real-time sensorimotor interactions with the environment \cite{liu2025embodied}. This kind of adaptive behavior inherently demands continuous inference under uncertainty.

Bayesian statistics offers a principled and coherent probabilistic framework to meet this challenge, encoding knowledge as probability distributions and refining beliefs through evidence-based updates \cite{ghahramani2015probabilistic}. As such, the core computational processes underlying embodied intelligence - including perception \cite{marton2013bayesian,feng2022bayesian}, action selection \cite{nazarczuk2024closed,andriella2025bayesian}, learning \cite{nguyen2023robot}, verification \cite{zhao2024bayesian}, and even higher-level cognition \cite{safron2021radically} - can be effectively understood and modeled as forms of Bayesian inference \cite{bellot2004bayesian}.

Despite the deep conceptual connection between Bayesian methods and embodied AI - as well as the theoretical elegance of the Bayesian framework - Bayesianism has not played a central role in the development of modern embodied intelligence. This paper explores the reasons behind this, through the lens of \textbf{search} and \textbf{learning} - two foundational themes in contemporary AI research, as emphasized in Rich Sutton's ``The Bitter Lesson" \cite{sutton2019bitter}. At the same time, our analysis shows that current embodied intelligence systems remain largely confined to closed physical-world environments, and it highlights the potential for Bayesian methods to play a pivotal role in extending these systems toward truly open physical-world embodied intelligence.

The remainder of this paper is organized as follows. In Section \ref{sec:search_learning}, we revisit Rich Sutton's ``The Bitter Lesson". Section \ref{sec:ei} summarizes current common practices in embodied intelligence. In Section \ref{sec:connection}, we examine the conceptual and methodological connections between Bayesian and embodied intelligence. Section \ref{sec:how} discusses how Bayesian approaches might help shape the future of embodied intelligence. Finally, we conclude the paper in Section \ref{sec:conclusion}.
\section{Search and Learning: Two Foundational Themes in Modern AI}
\label{sec:search_learning}
In ``The Bitter Lesson,” Rich Sutton highlights that \textbf{search} and \textbf{learning} represent general-purpose methods capable of scaling with increased computation to drive major breakthroughs in artificial intelligence \cite{sutton2019bitter}.

In Sutton’s context, \textbf{search} refers to algorithms that systematically explore vast spaces of possible solutions to a given problem. This might involve tuning neural network parameters, simulating different strategies in a game, or optimizing sequences of actions. The success of these methods often hinges on efficiently navigating immense search spaces - something that becomes increasingly feasible as computational power grows.
\textbf{Learning} involves training models on data to improve their ability to perform tasks. This mainly includes supervised learning (learning from labeled examples) \cite{hastie2008overview} and reinforcement learning (learning through trial and error with feedback from the environment) \cite{sutton1998reinforcement,luo2024precise,franccois2018introduction}.

Sutton’s ``bitter" insight is that systems designed by humans - though often successful in the early stages - tend to plateau in performance. In contrast, systems built upon scalable, general-purpose methods such as search and learning continue to improve with increased computational resources. The ``bitterness" stems from the fact that researchers are often inclined to embed domain-specific knowledge into their systems, an approach that can ultimately impede long-term progress. Classic examples include chess and computer vision .
Early chess engines were grounded in expert heuristics and handcrafted knowledge. By contrast, AlphaZero, leveraging deep learning and large-scale search, learned superhuman strategies from first principles and surpassed traditional engines \cite{silver2018general}. Similarly, conventional computer vision systems relied heavily on manually engineered features, whereas modern deep learning models learn hierarchical representations directly from data, achieving significantly superior performance \cite{krizhevsky2012imagenet,szegedy2015going,he2016deep}.

In essence, ``The Bitter Lesson" teaches that the most powerful AI systems will be those that harness increasing computational capacity through general-purpose search and learning, rather than those that rely heavily on human knowledge or specialized designs.
\section{Current Common Practices for Embodied Intelligence}
\label{sec:ei}
In this section, we present embodied intelligence methodologies adopted by major practices in the present period. By ``the present period" we mean the time since the emergence of ChatGPT and the rise of large language models (LLMs) as a widely discussed topic in both academia and industry.

Current mainstream approaches to embodied intelligence are built upon recent advancements in AI foundation models, such as pre-trained LLMs and vision-language models (VLMs). Broadly speaking, these foundation models endow embodied agents or robots - such as humanoids - with rich \emph{prior knowledge} about the world. Unlike predefined, human-engineered logic, this \emph{prior knowledge} is acquired through large-scale data-driven learning, such as training on internet-scale text corpora for LLMs. A robot equipped with one or more foundation models - such as Deepseek R1 \cite{guo2025deepseek}, GPT-4o \cite{hurst2024gpt}, Qwen2.5-VL \cite{bai2025qwen2}, or CLIP \cite{radford2021learning} - effectively possesses a cognitive core akin to a brain. Its capacity to interpret both visual and textual information is significantly enhanced, enabling it to perform complex tasks by understanding context, recognizing objects, and following instructions in a manner more comparable to human interaction.

However, foundation models alone do not satisfy the full range of requirements for embodied intelligence systems. The \emph{prior knowledge} encoded in these models is static and coarse-grained, which is insufficient for guiding and controling a robot to perform precise, fine-grained actions in dynamic environments - particularly within specialized domains whose relevant information was not covered in the training data of these models.

From the perspective of model and algorithm design, there are currently two main approaches to addressing the above issue. The first approach is to treat the pre-trained foundation model as a building block within a broader system and augment it with additional modules - such as memories \cite{xie2024embodied}; atomic skill models e.g., for planning \cite{hu2023enabling,dasgupta2023collaborating,zhang2023building,song2023llm}, perception \cite{fung2025embodied}, sensorimotor control \cite{mon2025embodied}, and navigation \cite{wang2025navrag}. The second is to collect domain-specific data and either fine-tune the foundation models or train new models based on them, enabling the resulting model to map perceptual and instruction inputs to actions in an end-to-end manner. We refer to the first category as Foundation Model-powered Systems (FMS), and the second as End-to-End modeling (E2E). Examples of FMS methods include \cite{hu2023enabling,dasgupta2023collaborating,zhang2023building,song2023llm}, and classic E2E approaches include GATO \cite{reed2022generalist}, the RT series \cite{brohan2022rt,zitkovich2023rt,vuong2023open}, the $\pi$ series \cite{black2410pi0,intelligence2025pi_}, and others e.g., OpenVLA \cite{kim2024openvla}.
\subsection{The Learning and Search Operations Embedded in Current Practices}
The core philosophy underlying the aforementioned approaches can be summarized in one phrase: data-driven \textbf{learning}. In FMS, certain parts of the system - whether a building block of the system or an interface between two building blocks - are learned from data. In E2E type approaches, a subset of neural network weights of a foundation model or the entire model is trained directly from data. Key relevant research topics include: how to collect and build up a high-quality dataset for learning a building block, fine-tuning or training from scratch a foundation model; how to design effective model architectures; how to align or fuse multi-modal data in one model; how to define suitable loss functions; how to design modular building blocks and compose them into a system; and how to determine which building blocks should be learned from data versus specified \emph{a priori} by humans.

In such a data-driven \textbf{learning} paradigm, for E2E-type methods, the \textbf{search} operation is not explicitly exposed but embedded in the model training or fine-tuning process, aiming to find parameter values that minimize a loss function tailored to the task. Stochastic gradient descent methods, such as Adam \cite{kingma2014adam}, are among the most commonly used optimization approaches to train deep neural network models. FMS-type approaches may incorporate explicit search mechanisms for planning, akin to the Monte Carlo Tree Search (MCTS) \cite{browne2012survey,swiechowski2023monte} employed in systems like AlphaGo \cite{silver2016mastering} and AlphaZero \cite{silver2018general}.
\section{Connections between Bayesian and Embodied Intelligence}
\label{sec:connection}
\subsection{From a deep philosophical perspective, Bayesianism and embodied intelligence are closely connected}
Bayesianism interprets probability as a quantification of subjective belief and emphasizes the dynamic updating of knowledge through evidence (from prior to posterior). At its core, it acknowledges the incompleteness of cognition and seeks to approximate truth through iterative updates. In another word, Bayesianism views an agent's ``mind" or intelligence as a process of belief updating. In a similar spirit, Embodied intelligence argues that intelligence is an emergent phenomenon arising from continuous interactions of body, environment, and brain, where the body is considered a subject of cognition rather than a passive object. Thus, both embodied intelligence and Bayesianism share a homologous learning mechanism: they view cognition/intelligence as dependent on dynamic interaction rather than static data. This aligns with the paradigm of emergent intelligence - where intelligence is neither innate nor merely acquired, but arises from the ongoing dialogue between the agent (including its body) and the world. See Table \ref{tab:bayes_vs_embodied} for a comparative analysis between intelligence implemented using Bayesian principles - referred to here as Bayesian intelligence - and embodied intelligence.
\begin{table}[t]
    \centering
        \caption{Bayesian Intelligence vs. Embodied Intelligence}
    \begin{tabular}{c| c| c}
    \hline
     Perspective   &   Bayesian Intelligence & Embodied Intelligence\\
   \hline
     Definition & Rational decision-making under uncertainty & Intelligence grounded in an agent’s physical\\
                & using probabilistic inference & interaction with the real world\\
   \hline
     Emphasis & Internal models of the world  & Situatedness, sensorimotor coupling,  \\
              & (belief updating, latent variables) & action-perception loops \\
   \hline
    \end{tabular}
    \label{tab:bayes_vs_embodied}
\end{table}

A natural question arises: Can embodied intelligence be realized through Bayesian mechanisms, and if so, how? Embodied intelligence holds that cognitive capabilities fundamentally emerge from - and are shaped by - an agent's real-time sensorimotor interactions with its environment. Such adaptive behavior inherently demands continuous inference under uncertainty. In principle, Bayesian statistics provide a rigorous probabilistic framework for meeting this challenge by representing knowledge as probability distributions and dynamically updating beliefs in response to new evidence. The core computational processes underlying embodied intelligence - including perception, action selection, learning, and even higher-level cognition - can be effectively interpreted and modeled as forms of Bayesian inference. But in reality, the mainstream frameworks for implementing modern embodied intelligence are not based on Bayesian methods. Why is this the case? In ``The Bitter Lesson", Rich Sutton emphasizes that modern AI should embrace scalable computational paradigms - centered on \textbf{learning} and \textbf{search} - that can fully leverage the ever-increasing availability of data and computational power. To understand why Bayesian approaches have not been widely adopted in contemporary embodied intelligence, we reexamine them from the perspectives of \textbf{learning} and \textbf{search}.
\subsection{From a learning and search perspective, there remains a significant gap between Bayesianism and current practices in embodied intelligence}
\label{sec:gap}
\textbf{Learning} can be framed as Bayesian inference over hidden variables or model parameters - for example, inferring the posterior distribution over weights or latent causes. Compared with other learning paradigms, Bayesian approaches offer a more principled way of handling uncertainty and incorporating prior knowledge. In addition, they provide a theoretically grounded framework for continual incremental learning, where data arrives sequentially, one item at a time. In other words, the Bayesian toolkit includes powerful methods, e.g., sequential monte carlo methods, capable of learning from a dataset in a sequential manner, e.g., by accessing each data point only once \cite{liu2023robust,liu2020sequential,xu2023bayesian,kumar2021bayesian,sadeghi2017embodied,zhang2021variational}.

In contrast, modern embodied intelligence frameworks built on deep neural networks are typically trained in batch mode, where each data item is accessed multiple times during training. Conceptually, the Bayesian learning paradigm is more human-like. In human learning, there is no clear distinction between a training phase and an inference phase—we learn incrementally and continuously. This raises a natural question: why are contemporary embodied intelligence systems predominantly based on batch-mode learning rather than Bayesian-style continual incremental learning?

Sutton’s ``The Bitter Lesson" offers an important clue. He argues that long-term progress in AI is driven by scalable methods rooted in search and learning, rather than approaches that rely heavily on human knowledge or hand-crafted structure. As discussed in Section \ref{sec:ei}, prevailing practices in embodied intelligence embrace scalable, assumption-light learning paradigms, such as self-supervised learning from large-scale text, image, and video data. A dominant architecture enabling such self-supervised learning is the transformer \cite{vaswani2017attention,wang2024scaling}.

In contrast, Bayesian learning methods often depend on structured priors or explicit model assumptions - such as those found in graphical models - which can hinder scalability. This reliance on structure contrasts with Sutton's call for data-driven, assumption-light approaches that better exploit the benefits of scale. In Table \ref{tab:Synthesis}, we provide a contrast for Bayesian intelligence and Sutton's ``bitter" preference.

\begin{table}[t]
    \small
    \centering
        \caption{Bayesian Intelligence vs. Sutton's ``Bitter" Preference}
    \begin{tabular}{c| c| c}
    \hline
     Feature   &   Bayesian Intelligence & Sutton's ``Bitter" Preference\\
   \hline
     Model Dependence & High  & Low model dependence\\
                & (explicit probabilistic models) & \\
   \hline
     Human Knowledge Injection & Frequent   & Minimal hand-crafting  \\
             & (priors, likelihoods, structures) &    \\
   \hline
   Learning Scalability & Limited by  & Scalable methods favored \\
              &  computation/inference complexity &  \\
   \hline
   Search Method & Internal, guided by  & External, such as\\
                 & belief updates & Monte Carlo Tree search \\
    \hline
    \end{tabular}
    \label{tab:Synthesis}
\end{table}
\section{How might Bayesian approaches help shape the future of embodied intelligence?}
\label{sec:how}
As mentioned above, modern embodied intelligence systems, particularly those built on deep learning and large pre-trained models, have embraced data-driven, assumption-light approaches that scale well with computation and massive datasets, and this paradigm in spirit aligns with Sutton’s ``bitter" preferences.

As a data-driven paradigm, its effectiveness hinges on access to a substantial amount of pre-acquired training data that closely aligns with the conditions of the deployment environment.

There are two main strategies to mitigate the challenges posed by data scarcity. The first is to collect human demonstration data \cite{fu2024mobile,zhao2023learning}, a practice commonly seen in the domain of autonomous driving, where every human driver effectively acts as a ``teacher" for the AI system. However, this approach does not readily generalize to other domains. Consider, for example, a robot designed for household services. We expect it to autonomously carry out a wide variety of tasks—such as storage, cleaning, laundry, cooking, and using various tools - across highly diverse household settings. Yet the layout of each home is different, and so are the objects within, including the types of windows, doors, and tools. As a result, training a robot to perform a wide range of general-purpose service tasks across many different household scenarios would likely require significantly more data than is needed for autonomous driving systems. At present, however, we lack a scalable method for collecting the equivalent of ``human driving" data for this kind of household service–oriented ``autonomous driving system."

An alternative approach is to resort to simulation - that is, to construct a digital counterpart of the physical world. If this simulated environment sufficiently resembles the real world, it can generate as much training data as needed for embodied intelligence models. Building such a simulation system is essentially equivalent to constructing a world model. However, as we know, the real world is immensely complex, and approximating its full dynamics through a comprehensive world model remains a virtually impossible task - at least for now.

Current large-scale pre-trained models - such as LLMs and vision-language models (VLMs) - can at best be considered coarse-grained approximations of a world model. They fall far short of supporting embodied intelligence in the rich, dynamic, and three-dimensional physical world. In practice, today's embodied AI systems are typically confined to a predefined scope of operation. This working scope corresponds to a simplified, constrained physical environment - small enough for us to model and simulate with reasonable fidelity. Naturally, such an environment has boundaries, and the resulting embodied AI system is only effective within those boundaries. In other words, \emph{\textbf{these systems operate in closed physical worlds, rather than truly open ones}}.

From an endgame perspective, however, embodied intelligence should be capable of functioning in open physical environments. In this broader setting, all the knowledge and skills acquired in closed-world settings can be regarded as prior knowledge - to borrow terminology from Bayesian inference. Upon entering an open world, the embodied agent engages in real-time sensorimotor interactions with its environment and must continuously adapt its behavior. This kind of adaptive behavior fundamentally requires ongoing inference under uncertainty. The core computational processes underlying embodied intelligence—including perception, action selection, learning, and even higher-level cognition—can be effectively understood as forms of Bayesian inference. This suggests that an embodied intelligence system designed to operate in an open physical world can be framed as a hierarchical Bayesian inference engine.

In addition, a variety of ready-to-use Bayesian methods have been developed for derivative-free global optimization of complex systems, particularly where conventional gradient-based approaches are inapplicable. For example, Bayesian optimization has been widely applied to tasks such as neural network architecture search and automated machine learning (AutoML) \cite{shahriari2015taking, malkomes2016bayesian}, both of which are closely tied to the broader pursuit of artificial general intelligence \cite{liu2018very}. Embodied intelligence–driven robotic systems represent a class of complex systems that can benefit from such Bayesian computational tools. For instance, Sequential Monte Carlo (SMC) optimization methods \cite{liu2017posterior, liu2016particle} have been employed for the automated co-design of soft hand morphology and control strategies for grasping, leveraging their inherent support for parallel computation \cite{deimel2017automated}. The global system-level optimization of morphology, action, perception, and learning has become an emerging trend in the development of embodied robotics \cite{liu2025embodied}. Beyond optimization, Bayesian techniques have also been used to bridge the sim-to-real gap \cite{muratore2021data,antonova2020sim2real,}, enhance the robustness of reinforcement or imitation learning \cite{derman2020bayesian,wei2023bayesian,brown2020bayesian,zhao2023context}, and enable multi-fidelity data fusion \cite{li2020multi, mikkola2023multi, liu2020harnessing, meng2021multi}, among other applications.

Importantly, the reliance of Bayesian methods on structured model assumptions can be relaxed—for example, by operating over an ensemble or a set of candidate models rather than committing to a single fixed model—thus improving flexibility and generalization in real-world scenarios \cite{liu2023robust}.

To summarize, the Bayesian framework offers principled tools for uncertainty quantification, probabilistic reasoning, and incremental learning - all of which are essential for operating in open, dynamic, and partially observable environments. These capabilities position Bayesian methods as a promising foundation for enabling embodied intelligence systems to function effectively in the truly open physical world.
\section{Concluding Remarks}
\label{sec:conclusion}
This paper explores the underexamined yet conceptually rich connection between Bayesian statistics and embodied intelligence. Grounded in the foundational perspectives of search and learning, it sheds light on why Bayesian approaches have been largely absent from mainstream embodied intelligence frameworks (see Subsection \ref{sec:gap}). Furthermore, it highlights that current practices in embodied intelligence remain confined to closed physical-world settings, and argues that enabling embodied systems to operate in truly open physical-worlds may require harnessing the distinctive strengths of Bayesian methods (see Section \ref{sec:how}).

The focus of this paper is on fundamental theoretical discussions, without delving into specific models or algorithmic details. It is hoped that this effort will engage a broader audience in foundational thinking, thereby stimulating the development of new models and algorithms at the intersection of Bayesian methods and embodied intelligence, as well as the design of novel hardware architectures tailored for Bayesian computation. With such advances, truly embodied intelligence systems - capable of perception, locomotion, and manipulation in dynamic, open physical worlds - may soon become a tangible reality.

The analytical approach adopted in this paper - namely, analyzing from the perspectives of \textbf{Search} and \textbf{Learning}—is not only applicable to explaining why Bayesian methods have not become the mainstream computational paradigm in contemporary AI, but also equally applicable to understanding why other computational paradigms, though popular within certain communities, have likewise failed to become dominant in today’s AI landscape.
\bibliography{ref}
\bibliographystyle{rlc}
\section*{Biography}
Bin Liu holds a Ph.D. from the Chinese Academy of Sciences. Bayesian inference and embodied intelligence have been the two main research themes throughout his academic career. His exposure to Bayesian computation began during his PhD studies while working on state estimation problems in dynamical systems. Later, at Duke University, he collaborated with statisticians Jim Berger and Merlise Clyde, as well as Tom Loredo, an astronomer from Cornell University - who was also an expert in Bayesian computation. During this period, he also participated in the Sequential Monte Carlo (SMC) program held at Statistical and Applied Mathematical Sciences Institute, a US national research institute. After many years of research on Bayesian methods, he came to realize that, despite their theoretical elegance, Bayesian approaches alone are far from sufficient to tackle the complexities of real-world problems. This realization led him to embrace deep learning, reinforcement learning, and especially large-scale pre-trained models, in an effort to integrate the strengths of diverse methodologies and build a framework for embodied learning and decision-making that can truly function in dynamic and open physical environments. He is currently the Chief Robot Expert at E-Surfing Digital Life Technology Co., Ltd., and a Level 2 Group Chief Expert at China Telecom. He has established a robotics laboratory focused on home environments and leads a team dedicated to developing advanced embodied intelligence technologies and general-purpose robotic systems.
\end{document}